\documentclass[]{article}
\usepackage{proceed2e}
\usepackage{times} 
\usepackage{subscript} 
\usepackage[geometry]{ifsym} 
\usepackage{color} 
\usepackage[utf8]{inputenc} 
\usepackage{amsmath,amssymb,amsthm} 
\usepackage{graphicx} 
\usepackage{subfigure} 
\usepackage{booktabs} 
\usepackage{multirow} 
\usepackage{multicol}
\usepackage[round]{natbib}
\usepackage{url}
\usepackage{caption}
\usepackage{paralist}
\usepackage[lined, ruled, boxed, linesnumbered, commentsnumbered]{algorithm2e}
\usepackage{titlesec} 
\titlespacing\section{0pt}{12pt plus 4pt minus 2pt}{0pt plus 2pt minus 2pt}
\titlespacing\subsection{0pt}{12pt plus 4pt minus 2pt}{0pt plus 2pt minus 2pt}
\titlespacing\subsubsection{0pt}{12pt plus 4pt minus 2pt}{0pt plus 2pt minus 2pt}

\newcommand{\argmin}[1]{\underset{#1}{\operatorname{arg}\!\operatorname{min}}\;}

\newcommand{\Do}[1]{do(#1)}
\newcommand{\man}[1]{man(#1)}

\newtheorem{theorem}{Theorem}

\newtheorem{definition}[theorem]{Definition}

\title{Visual Causal Feature Learning}

\author{ {\bf Krzysztof Chalupka} \\
Computation and Neural Systems\\
California Institute of Technology\\
Pasadena, CA, USA\\
\And
{\bf Pietro Perona}  \\
Electrical Engineering\\
California Institute of Technology\\
Pasadena, CA, USA\\
\And
{\bf Frederick Eberhardt}   \\
Humanities and Social Sciences\\
California Institute of Technology\\
Pasadena, CA, USA\\
}

\begin{document}

\maketitle

\begin{abstract}
We provide a rigorous definition of the \emph{visual cause} of a behavior that is broadly applicable to the visually driven behavior in humans, animals, neurons, robots and other perceiving systems. Our framework generalizes standard accounts of causal learning to settings in which the causal variables need to be constructed from micro-variables. We prove the Causal Coarsening Theorem, which allows us to gain causal knowledge from observational data with minimal experimental effort. The theorem provides a connection to standard inference techniques in machine learning that identify features of an image that \emph{correlate} with, but may not \emph{cause}, the target behavior. Finally, we propose an active learning scheme to learn a manipulator function that performs optimal manipulations on the image to automatically identify the visual cause of a target behavior. We illustrate our inference and learning algorithms in experiments based on both synthetic and real data. 
\end{abstract}

\section{INTRODUCTION}
Visual perception is an important trigger of human and animal behavior.
The visual cause of a behavior can be easy to define, say, when a traffic light turns green, or quite subtle: apparently it is the increased symmetry of features that leads people to judge faces more attractive than others~\citep{Grammer1994}. Significant scientific and economic effort is focused on visual causes in advertising, entertainment, communication, design, medicine, robotics and the study of human and animal cognition.
Visual causes profoundly influence our daily activity, yet our understanding of what constitutes a visual cause lacks a theoretical basis. In practice, it is well-known that images are composed of millions of variables (the pixels) but it is functions of the pixels (often called `features') that have meaning, rather than the pixels themselves.

We present a theoretical framework and inference algorithms for visual causes in images. A visual cause is defined (more formally below) as a function (or \emph{feature}) of raw image pixels that has a \emph{causal effect} on the target behavior of a perceiving system of interest. We present three advances:
\begin{compactitem}
\item We provide a definition of the visual cause of a target behavior as a macro-variable that is constructed from the micro-variables (pixels) that make up the image space. The visual cause is distinguished from other macro-variables in that it contains all the causal information about the target behavior that is available in the image. We place the visual cause within the standard framework of causal graphical models~\citep{Spirtes2000,Pearl2009}, thereby contributing to an account of how to construct causal variables.
\item We prove the Causal Coarsening Theorem (CCT), which shows how observational data can be used to learn the visual cause with minimal experimental effort. It connects the present results to standard classification tasks in machine learning.
\item We describe a method to learn the manipulator function, which automatically performs perceptually optimal manipulations on the visual causes.
\end{compactitem}

We illustrate our ideas using synthetic and real-data experiments. Python code that implements our algorithms, as well as reproduces some of the experimental results, is available online at \url{http://vision.caltech.edu/~kchalupk/code.html}. 

We chose to develop the theory within the context of \textit{visual} causes as this setting makes the definitions most intuitive and is itself of significant practical interest. However, the framework and results can be equally well applied to extract causal information from any aggregate of micro-variables on which manipulations are possible. Examples include auditory, olfactory and other sensory stimuli; high-dimensional neural recordings; market data in finance; consumer data in marketing. There, causal feature learning is both of theoretical (``What is the cause?'') and practical (``Can we automatically manipulate it?'') importance.

\subsection{PREVIOUS WORK}
Our framework extends the theory of causal graphical models~\citep{Spirtes2000, Pearl2009} to a setting in which the input data consists of raw pixel (or other micro-variable) data. In contrast to the standard setting, in which the macro-variables in the statistical dataset already specify the candidate causal relata, the causal variables in our setting have to be constructed from the micro-variables they supervene on, before any causal relations can be established. We emphasize the difference between our method of causal feature \emph{learning} and methods for causal feature \emph{selection}~\citep{Guyon2007a, Pellet2008a}. The latter choose the best (under some causal criterion) features from a restricted set of plausible macro-variable candidates. In contrast, our framework efficiently searches the whole space of all the possible macro-variables that can be constructed from an image. 

Our approach derives its theoretical underpinnings from computational mechanics \citep{Shalizi2001,Shalizi2001a}, but supports a more explicitly causal interpretation by incorporating the possibility of confounding and interventions. Since we allow for unmeasured common causes of the features in the image and the target behavior, we have to distinguish between the plain conditional probability distribution of the target behavior ($T$) given the (observed) image ($I$) and the distribution of the target behavior given that the observed image was manipulated (i.e.\ $P(T|I)$ vs.\ $P(T|do(I))$). \citet{Hoel2013}, who develop a similar model to investigate the relationship between causal micro- and macro-variables, avoid this distinction by assuming that all their data was generated from what in our setting would be the manipulated distribution $P(T|do(I))$. We take the distinction between interventional and observational distributions to be one of the key features of a causal analysis.
The extant literature on causal learning from image or video data does not generally consider the aggregation from pixel variables into causal macro-variables, but instead starts from annotated or pre-defined features of the image (see e.g.\ \citet{Fire2013a,Fire2013b}).

\subsection{CAUSAL FEATURE LEARNING: AN EXAMPLE}
\label{sec:example}
\begin{figure}
\centering
\includegraphics{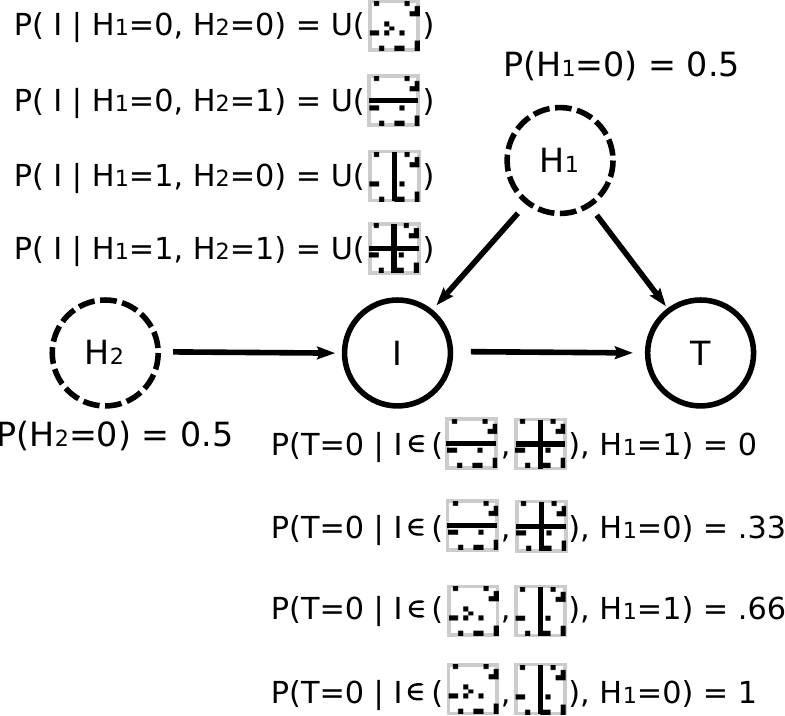}
\caption{Our case study generative model. Two binary hidden (non-visual) variables $H_1$ and $H_2$ toss unbiased coins. The content of the image $I$ depends on these variables as follows. If $H_1=H_2=0$, $I$ is chosen uniformly at random from all the images containing no v-bars and no h-bars. If $H_1=0$ and $H_2=1$, $I$ is chosen uniformly at random from all images containing at least one h-bar but no v-bars. If $H_1=1$ and $H_2=0$, $I$ is chosen uniformly at random from all the images containing at least one v-bar but no h-bars. Finally, if $H_1=H_2=1$, $I$ is chosen from images containing at least one v-bar and at least one h-bar. The distribution of the binary behavior $T$ depends only on the presence of an h-bar in $I$ and the value of $H_1$. In observational studies, $H_1=1$ iff $I$ contains a v-bar. However, a \emph{manipulation} of any specific image $I=i$ that introduces a v-bar (without changing $H_1$) will in general not change the probability of $T$ occurring. Thus, $T$ does \emph{not} depend causally on the presence of v-bars in $I$.}
\label{fig:grating_data}
\end{figure}

Fig.~\ref{fig:grating_data} presents a paradigmatic case study in visual causal feature learning, which we will use as a running example. The contents of an image $I$ are caused by external, non-visual binary hidden variables $H_1$ and $H_2$ such that if $H_1$ is on, $I$ contains a vertical bar (v-bar\footnote{We take a v-bar (h-bar) to consist of a complete column (row) of black pixels.}) at a random position, and if $H_2$ is on, $I$ contains a horizontal bar (h-bar) at a random position. A target behavior $T\in\{0,1\}$ is caused by $H_1$ and $I$, such that $T=1$ is more likely whenever $H_1=1$ and whenever the image contains an {h-bar}. 

We deliberately constructed this example such that the visual cause is clearly identifiable: manipulating the presence of an h-bar in the image will influence the distribution of $T$. Thus, we can call the following function $C\colon \mathcal{I}\rightarrow \{0,1\}$ the \emph{causal feature} of $I$ or the \emph{visual cause} of $T$:
\[
C(I) = \left\{\begin{array}{l} 1\quad\text{if }I\text{ contains an h-bar}\\
                                0\quad\text{otherwise}.
              \end{array}\right.
\]
The presence of a v-bar, on the other hand, is not a causal feature. Manipulating the presence of a v-bar in the image has no effect on $H_1$ or $T$. Still, the presence of a v-bar is as strongly correlated with the value of $T$ (via the common cause $H_1$) as the presence of an h-bar is. We will call the following function $S\colon \mathcal{I}\rightarrow \{0,1\}$ the \emph{spurious correlate} of $T$ in $I$:
\[
S(I) = \left\{\begin{array}{l} 1\quad\text{if }I\text{ contains a v-bar}\\
                                0\quad\text{otherwise}.
              \end{array}\right.
\]

Both the presence of h-bars and the presence of v-bars are good individual (and even better joint) predictors of the target variable, but only one of them is a cause. Identifying the visual cause from the image thus requires the ability to distinguish among the correlates of the target variables those that are actually causal, even if the non-causal correlates are (possibly more strongly) correlated with the target.

While the values of $S$ and $C$ in our example stand in a bijective correspondence to the values of $H_1$ and $H_2$, respectively, this is only to keep the illustration simple.
In general, the visual cause and the spurious correlate can be probabilistic functions of any number of (not necessarily independent) hidden variables, and can share the same hidden causes.

\section{A THEORY OF VISUAL CAUSAL FEATURES}
In our example the identification of the visual cause with the presence of an h-bar is intuitively obvious,
as the model is constructed to have an easily describable visual cause. But the example does not provide a theoretical account of what it takes to be a visual cause in the general case when we do not know what the causally relevant pixel configurations are. In this section, we provide a general account of how the visual cause is related to the pixel data.

\subsection{VISUAL CAUSES AS MACRO-VARIABLES}
A visual cause is a high-level random variable that is a function (or feature) of the image, which in turn is defined by the random micro-variables that determine the pixel values. The functional relation between the image and the visual cause is, in general, surjective, though in principle it could be bijective. While we are interested in identifying the visual causes of a target behavior, the functional relation between the image pixels and the visual cause should not itself be interpreted as causal. Pixels do not \emph{cause} the features of an image, they \emph{constitute} them, just as the atoms of a table constitute the table (and its features). The difference between the causal and the constitutive relation is that the former requires the possibility of independent manipulation (at least to some extent), whereas by definition one cannot manipulate the visual cause without manipulating the image pixels. 

The probability distribution over the visual cause is induced by the probability distribution over the pixels in the image and the functional mapping from the image to the visual cause. But since a visual cause stands in a constitutive relation with the image, we cannot without further explanation describe interventions on the visual cause in terms of the standard $do$-operation~\citep{Pearl2009}. Our goal will be to define a macro-variable $C$, which contains all the causal information available in an image about a given behavior $T$, and define its manipulation.
To make the problem approachable, we introduce two (natural) assumptions about the causal relation between the image and the behavior: (i)~The value of the target behavior $T$ is determined subsequently to the image in time, and (ii)~the variable $T$ is in no way represented in the image. These assumptions exclude the possibility that $T$ is a cause of features in the image or that $T$ can be seen as causing itself.

\subsection{GENERATIVE MODELS: FROM MICRO- TO MACRO-VARIABLES}
\label{sec:generativeModels}

\begin{figure}
\centering
\includegraphics{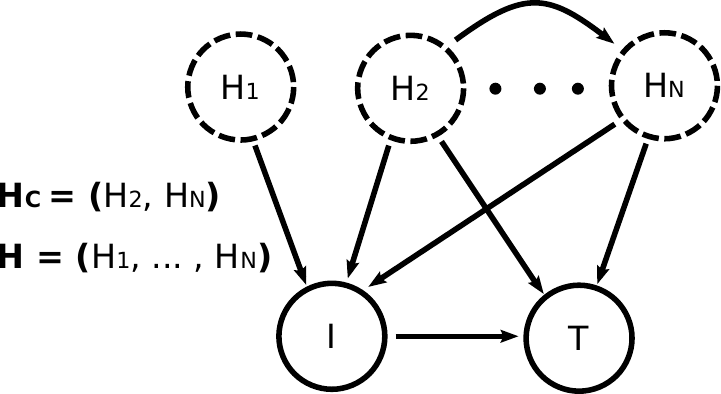}
\caption{A general model of visual causation. In our model each image $I$ is caused by a number of hidden non-visual variables $H_i$, which need not be independent. The image itself is the only observed cause of a target behavior $T$. In addition, a (not necessarily proper) subset of the hidden variables can be a cause of the target behavior. These confounders create visual ``spurious correlates'' of the behavior in $I$.}
\label{fig:genmodel}
\end{figure}

Let $T \in \{0,1\}$ represent a target behavior.\footnote{An extension of the framework to non-binary, discrete $T$ is easy but complicates the notation significantly. An extension to the continuous case is beyond the scope of this article.} Let $\mathcal{I}$ be a discrete space of all the images that can influence the target behavior (in our experiments in Section~\ref{sec:experiments}, $\mathcal{I}$ is the space of $n$-dimensional black-and-white images). We use the following generative model to describe the relation between the images and the target behavior: An image is generated by a finite set of unobserved discrete variables $H_1,\ldots, H_m$ (we write $\mathbf{H}$ for short). The target behavior is then determined by the image and possibly a subset of variables $\mathbf{H}_{c} \subseteq \mathbf{H}$ that are confounders of the image and the target behavior:
\begin{align}
P(T, I) &= \sum_{\mathbf{H}} P(T\mid I, \mathbf{H}) P(I \mid \mathbf{H}) P(\mathbf{H})\notag\\
        &= \sum_{\mathbf{H}} P(T\mid I, \mathbf{H}_{c}) P(I \mid \mathbf{H}) P(\mathbf{H}).\label{eq:genmodel}
\end{align} 
Independent noise that may contribute to the target behavior is marginalized and omitted for the sake of simplicity in the above equation. The noise term incorporates any hidden variables which influence the behavior but stand in no causal relation to the image. Such variables are not directly relevant to the problem. Fig.~\ref{fig:genmodel} shows this generative model. 

Under this model, we can define an \emph{observational partition} of the space of images $\mathcal{I}$ that groups images into classes that have the same conditional probability $P(T\mid I)$:

\begin{definition}[Observational Partition, Observational Class]
The \emph{observational partition} $\Pi_o(T, \mathcal{I})$ of the set $\mathcal{I}$ w.r.t. behavior $T$ is the partition induced by the equivalence relation $\sim$ such that $i \sim j$ if and only if $P(T\mid I=i) = P(T\mid I=j).$ We will denote it as $\Pi_o$ when the context is clear. A cell of an observational partition is called an \emph{observational class}.
\end{definition}

In standard classification tasks in machine learning, the observational partition is associated with class labels. In our case, two images that belong to the same cell of the observational partition assign equal \emph{predictive} probability to the target behavior. Thus, knowing the observational class of an image allows us to predict the value of $T$. However, the predictive probability assigned to an image does not tell us the \emph{causal} effect of the image on $T$. For example, a barometer is widely taken to be an excellent predictor of the weather. But changing the barometer needle does not cause an improvement of the weather. It is not a (visual or otherwise) cause of the weather. In contrast, seeing a particular barometer reading may well be a \emph{visual cause} of whether we pack an umbrella. 

Our notion of a visual cause depends on the ability to manipulate the image.

\begin{definition}[Visual Manipulation]
A \emph{visual manipulation} is the operation $\man{I=i}$ that changes (the pixels of) the image to image $i\in\mathcal{I}$, while not affecting any other variables (such as $\mathbf{H}$ or $T$). That is, the manipulated probability distribution of the generative model in Eq.~\eqref{eq:genmodel} is given by $P(T \mid \man{I=i}) = \sum_{\mathbf{H}_c} P(T\mid I=i, \mathbf{H}_{c}) P(\mathbf{H}_c).$

\end{definition}
The manipulation changes the values of the image pixels, but does not change the underlying ``world", represented in our model by the $H_i$ that generated the image. Formally, the manipulation is similar to the $do$-operator for standard causal models. However, we here reserve the $do$-operation for interventions on causal \emph{macro}-variables, such as the visual cause of $T$. We discuss the distinction in more detail below.

We can now define the \emph{causal partition} of the image space (with respect to the target behavior $T$) as:

\begin{definition}[Causal Partition, Causal Class]
The causal partition $\Pi_c(T, \mathcal{I})$ of the set $\mathcal{I}$ w.r.t. behavior $T$  is the partition induced by the equivalence relation $\sim$ defined on $\mathcal{I}$ such that $i \sim j$ if and only if $P(T \mid \man{I=i}) = P(T \mid \man{I=j})$ for $i,j \in \mathcal{I}$. 
When the image space and the target behavior are clear from the context, we will indicate the causal partition by $\Pi_c$. A cell of a causal partition is called a \emph{causal class}. 
\end{definition}

The underlying idea is that images are considered causally equivalent with respect to $T$ if they have the same causal effect on $T$. Given the causal partition of the image space, we can now define the visual cause of $T$:

\begin{definition}[Visual Cause]
The \emph{visual cause} $C$ of a target behavior $T$ is a random variable whose value stands in a bijective relation to the causal class of $I$. 
\end{definition}

The visual cause is thus a function over $\mathcal{I}$, whose values correspond to the post-manipulation distributions $C(i) = P(T\mid \man{I=i})$. We will write $C(i) = c$ to indicate that the causal class of image $i \in \mathcal{I}$ is $c$, or in other words, that in image $i$, the visual cause $C$ takes value $c$. Knowing $C$ allows us to predict the effects of a visual manipulation $P(T \mid  \man{I=i})$, as long as we have estimated $P(T \mid  \man{I=i^*_k})$ for one representative $i^*_k$ of each causal class $k$.

\subsection{THE CAUSAL COARSENING THEOREM}
\label{sec:cct}
Our main theorem relates the causal and observational partitions for a given $\mathcal{I}$ and $T$. It turns out that in general the causal partition is a coarsening of the observational partition. That is, the causal partition aligns with the observational partition, but the observational partition may subdivide some of the causal classes.

\begin{figure}
\centering
\includegraphics{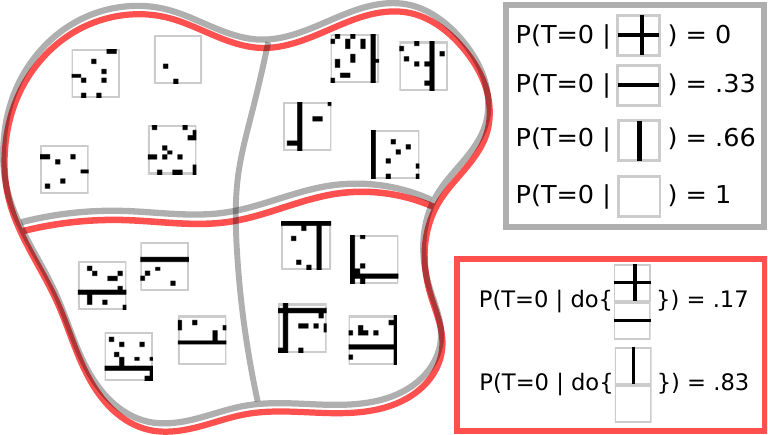}
\caption{The Causal Coarsening Theorem. The observational probabilities of $T$ given $I$ (gray frame) induce an observational partition on the space of all the images (left, observational partition in gray). The causal probabilities (red frame) induce a causal partition, indicated on the left in red.  The CCT allows us to expect that the causal partition is a coarsening of the observational partition. The observational and causal probabilities correspond to the generative model shown in Fig.~\ref{fig:grating_data}.}
\label{fig:cct}
\end{figure}

\begin{theorem}[Causal Coarsening] Among all the generative distributions of the form shown in Fig.~\ref{fig:genmodel} which induce a given observational partition $\Pi_o$, almost all induce a causal partition $\Pi_c$ that is a coarsening of the $\Pi_o$.~\label{thm:cct}
\end{theorem}
Throughout this article, we use ``almost all'' to mean ``all except for a subset of Lebesgue measure zero''. Fig.~\ref{fig:cct} illustrates the relation between the causal and the observational partition implied by the theorem. We note that the measure-zero subset where $\Pi_C$ does not coarsen $\Pi_O$ can indeed be non-empty. We provide such counter-examples in Appendix 7.

We prove the CCT in Appendix 6 using a technique that extends that of~\citet{Meek1995}: We show that (1) restricting the space of all the possible $P(T, H, I)$ to only the distributions compatible with a fixed observational partition puts a linear constraint on the distribution space; (2) requiring that the CCT be false puts a non-trivial polynomial constraint on this subspace, and finally, (3) it follows that the theorem holds for almost all distributions that agree with the given observational partition. The proof strategy indicates a close connection between the CCT and the faithfulness assumption \citep{Spirtes2000}. 

Two points are worth noting here: First, the CCT is interesting inasmuch as the visual causes of a behavior do not contain all the information in the image that predict the behavior. Such information, though not itself a cause of the behavior, can be informative about the state of other non-visual causes of the target behavior. Second, the CCT allows us to take any classification problem in which the data is divided into observational classes, and assume that the causal labels do not change within each observational class. This will help us develop efficient causal inference algorithms in Section~\ref{sec:inference}. 

\subsection{VISUAL CAUSES IN A CAUSAL MODEL CONSISTING OF MACRO-VARIABLES}
We can now simplify our generative model by omitting all the information in $I$ unrelated to behavior $T$. Assume that the observational partition $\Pi_o^T$ refines the causal partition~$\Pi_c^T$. Each of the causal classes $c_1, \cdots, c_K$ delineates a region in the image space $\mathcal{I}$ such that all the images belonging to that region induce the same $P(T \mid \man{I})$. Each of those regions---say, the k-th one---can be further partitioned into sub-regions $s^k_1, \cdots, s^k_{M_k}$ such that all the images in the m-th sub-region of the k-th causal region induce the same observational probability $P(T \mid I)$. By assumption, the observational partition has a finite number of classes, and we can arbitrarily order the observational classes within each causal class. Once such an ordering is fixed, we can assign an integer $m \in \{1, 2,\cdots, M_k\}$ to each image $i$ belonging to the k-th causal class such that $i$ belongs to the m-th observational class among the $M_k$ observational classes contained in $c_k$. By construction, this integer explains all the variation of the observational class within a given causal class. This suggests the following definition:

\begin{definition}[Spurious Correlate]
The \emph{spurious correlate} $S$ is a discrete random variable whose value differentiates between the observational classes contained in any causal class.
\end{definition}

The spurious correlate is a well-defined function on $\mathcal{I}$, whose value ranges between $1$ and $\max_k M_k$. Like $C$, the spurious correlate $S$ is a macro-variable constructed from the pixels that make up the image. $C$ and $S$ together contain all and only the visual information in $I$ relevant to $T$, but only $C$ contains the causal information:

\begin{theorem}[Complete Macro-variable Description] 
\label{thm:macro}
The following two statements hold for $C$ and $S$ as defined above:
\begin{enumerate}
\item $P(T\mid I) = P(T\mid C,S).$
\item Any other variable $X$ such that $P(T\mid I)=P(T\mid X)$ has Shannon entropy $H(X) \geq H(C,S)$.
\end{enumerate}
\end{theorem}

\begin{figure}
\centering
\includegraphics{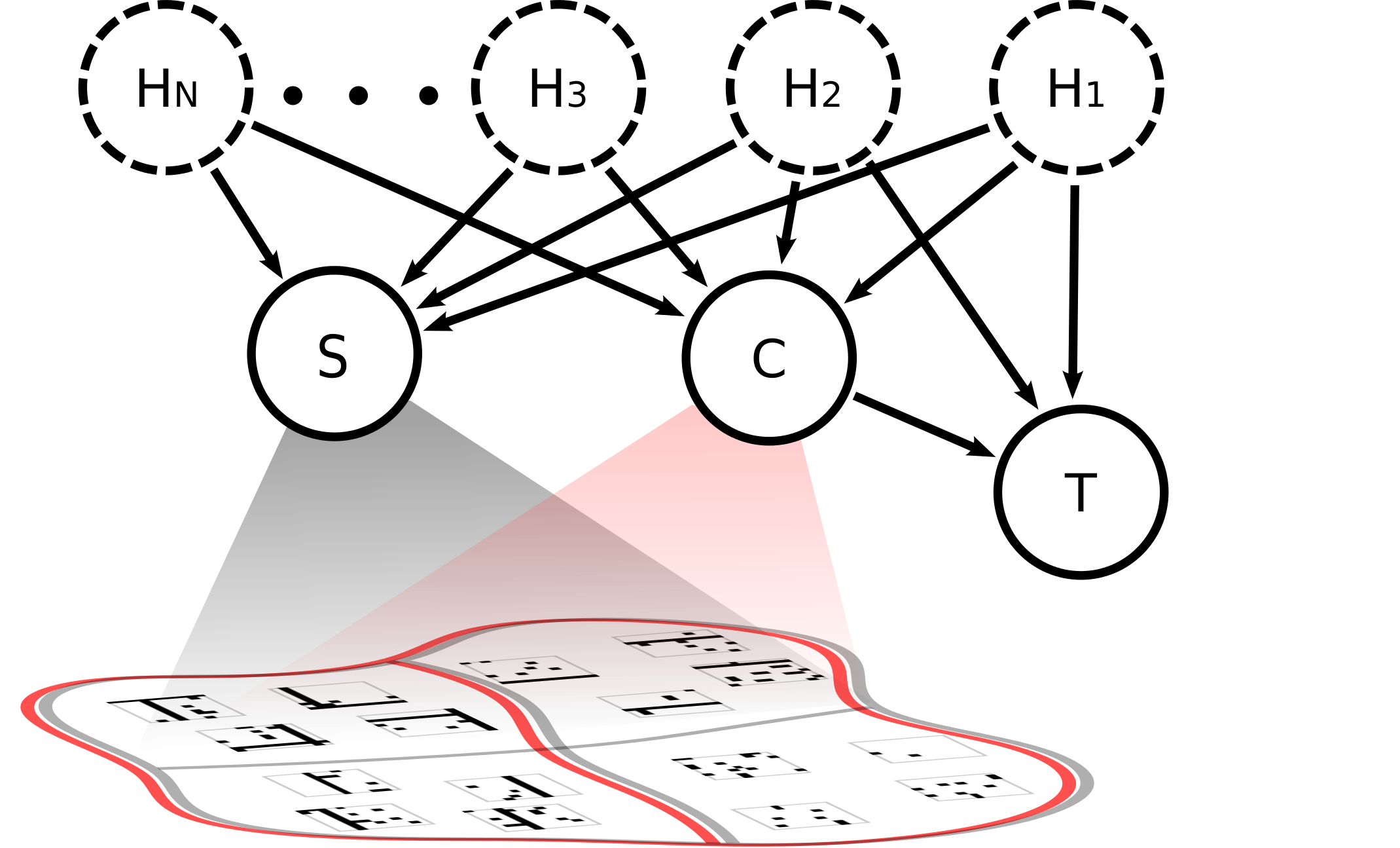}
\caption{A macro-variable model of visual causation. Using our theory of visual causation we can aggregate the information present in visual micro-variables (image pixels) into the visual cause $C$ and spurious correlate $S$. According to Theorem~\ref{thm:macro}, $C$ and $S$ contain all the information about $T$ available in $I$.}
\label{fig:macrovars}
\end{figure}

We prove the theorem in Appendix 8. It guarantees that $C$ and $S$ constitute the smallest-entropy macro-variables that encompass all the information about the relationship between $T$ and $I$. Fig.~\ref{fig:macrovars} shows the relationship between $C,S$ and $T$, the image space $\mathcal{I}$ and the observational and causal partitions schematically. $C$ is now a cause of $T$, $S$ correlates with $T$ due to the unobserved common causes $\mathbf{H}_C$, and any information irrelevant to $T$ is pushed into the independent noise variables (commonly not shown in graphical representations of structural equation models).\footnote{We note that $C$ may retain predictive information about $T$ that is not causal, i.e.\ it is not the case that all spurious correlations can be accounted for in $S$. See Appendix 9 for an example.}

The macro-variable model lends itself to the standard treatment of causal graphical models described in \citet{Pearl2009}. We can define interventions on the causal variables $\{C,S,T\}$ using the standard $do$-operation. The $do$-operator only sets the value of the intervened variable to the desired value, making it independent of its causes, but it does not (directly) affect the other variables in the system or the relationships between them (see the \emph{modularity assumption} in \citet{Pearl2009}). However, unlike the standard case where causal variables are separated in location (e.g.\ \emph{smoking} and \emph{lung cancer}), the causal variables in an image may involve the same pixels: $C$ may be the average brightness of the image, whereas $S$ may indicate the presence or absence of particular shapes in the image. An intervention on a causal variable using the $do$-operator thus requires that the underlying manipulation of the image respects the state of the other causal variables:

\begin{definition}[Causal Intervention on Macro-variables]
Given the set of macro-variables $\{C, S\}$ that take on values $\{c,s\}$ for an image $i \in \mathcal{I}$, an intervention $\Do{C=c'}$ on the macro-variable $C$ is given by the manipulation of the image $\man{I=i'}$ such that $C(i') = c'$ and $S(i') = s$. The intervention $\Do{S=s'}$ is defined analogously as the change of the underlying image that keeps the value of $C$ constant.\label{def:manipulation}
\end{definition}

In some cases it can be impossible to manipulate $C$ to a desired value without changing $S$. We do not take this to be a problem special to our case. In fact, in the standard macro-variable setting of causal analysis we would expect interventions to be much more restricted by physical  constraints than we are with our interventions in the image space. 

\section{CAUSAL FEATURE LEARNING: INFERENCE ALGORITHMS}
\label{sec:inference}
Given the theoretical specification of the concepts of interest in the previous section, we can now develop algorithms to learn $C$, the visual cause of a behavior.  In addition, knowledge of $C$ will allow us to specify a \emph{manipulator function}: a function that, given any image, can return a maximally similar image with the desired causal effect.

\begin{definition}[Manipulator Function]\label{def:manipulator}
Let $C$ be the causal variable of $T$ and $d$ a metric on $\mathcal{I}$. The \emph{manipulator function} of $C$ is a function $M_C \colon {\cal I}\times{\cal C}\to{\cal I}$ such that $M_C(i, k) = \arg\min_{\hat{\imath}\in C^{-1}(k)}d(i,\hat{\imath})$ for any $i\in\mathcal{I}, k\in\mathcal{C}.$ In case $d(i, .)$ has multiple minima, we group them together into one equivalence class and leave the choice of the representative to the manipulator function.
\end{definition}

The manipulator searches for an image closest to $I$ among all the images with the desired causal effect $k$. The meaning of ``closest'' depends on the metric $d$ and is discussed further in Section~\ref{sec:manipulator_practice} below. Note that the manipulator function can find candidates for the image manipulation underlying the desired causal manipulation $\Do{C=c},$ but it does not check whether other variables in the system (in particular, the spurious correlate) remain in fact unchanged. Using the closest possible image with desired causal effect is a heuristic approach to fulfilling that requirement. 

There are several reasons why we might want such a manipulator function:

\begin{compactitem}
\item If our goal is to perform causal manipulations on images, the manipulator function offers an automated solution.
\item A manipulator that uses a given $C$ and produces images with the desired causal effect provides strong evidence that $C$ is indeed the visual cause of the behavior.  
\item Using the manipulator function we can enrich our dataset with new datapoints, in hope of achieving better generalization on both the causal and predictive learning tasks.
\end{compactitem}

The problem of visual causal feature learning can now be posed as follows: Given an image space $\mathcal{I}$ and a metric $d$, learn $C$---the visual cause of $T$---and the manipulator $M_C$. 

\subsection{CAUSAL EFFECT PREDICTION}
A standard machine learning approach to learning the relation between $I$ and $T$ would be to take an \emph{observational dataset} $\mathcal{D}_{obs}=\{(i_k, P(T\mid i_k))\}_{k=1, \cdots, N}$ and learn a predictor $f$ whose training performance guarantees a low test error (so that $f(i^*) \approx P(T \mid i^*)$ for a test image $i^*$). In causal feature learning, low test error on observational data is insufficient; it is entirely possible that $\mathcal{D}$ contains spurious information useful in predicting test labels which is nevertheless not causal. That is, the prediction may be highly accurate for observational data, but completely inaccurate for a prediction of the effect of a manipulation of the image (recall the barometer example). However, we can use the CCT to obtain a causal dataset from the observational data, and then train a predictor on that dataset. Algorithm~\ref{alg:cslearn} uses this strategy to learn a function $C$ that, presented with any image $i\in\mathcal{I}$, returns $C(i)\approx P(T \mid  \man{I=i})$. We use a fixed neural network architecture to learn $C$, but any differentiable hypothesis class could be susbtituted instead. Differentiability of $C$ is necessary in Section~\ref{sec:manipulator_practice} in order to learn the  manipulator function.

\IncMargin{1.5em}
\begin{algorithm}[t!]
\caption{Causal Predictor Training}
\label{alg:cslearn}
\SetKwFunction{Train}{Train}
\SetKwData{Data}{data}\SetKwData{NN}{NN}
\SetKwInOut{Input}{input}
\SetKwInOut{Output}{output}

\Input{$\mathcal{D}_{obs} = {\{(i_1, p_1=p(T\mid i_1))},\cdots,$\\ \hspace{2mm}$(i_N, p_N=p(T\mid i_N)\}$ 
  -- observational data \\
       $\mathcal{P} = \{P_1, \cdots, P_M\}$ -- the set of observatio-\\
       nal classes (so that $\forall k, p_k \in \mathcal{P}, 1\leq k\leq N$)\\
       $\Train$ -- a neural net training algorithm}
\Output{$C\colon \mathcal{I}\to [0, 1]$ -- the causal variable}
\BlankLine
Pick $\{i_{k_1}, \cdots, i_{k_M}\}\subset\{i_1,\cdots, i_N\}$ 
    s.t. $p_{k_m}=P_m$\;\label{alg:cslearn_csldata}
Estimate $\hat{C}_m \leftarrow P(T \mid  \man{I=i_{k_m}})$ for each $m$\;\label{alg:cslearn_estimate}
For all $k$ let $\hat{C}(i_k) \leftarrow \hat{C}_m$ if $p_k = P_m$\;
$\mathcal{D}_{csl} \leftarrow \{(i_1, \hat{C}(i_1)), \cdots, (i_N, \hat{C}(i_N))\}$\;\label{alg:cslearn_coarsen}
$C \leftarrow \Train(\mathcal{D}_{csl})$\;\label{alg:cslearn_learn}
\end{algorithm}\DecMargin{1.5em}

In Step~\ref{alg:cslearn_csldata} the algorithm picks a representative member of each observational class. The CCT tells us that the causal partition coarsens the observational one. That is, in principle (ignoring sampling issues) it is sufficient to estimate $\hat{C}_m=P(T\mid \man{I=i_{k_m}})$ for just one image in an observational class $m$ in order to know that $P(T\mid \man{I=i})=\hat{C}_m$ for any other $i$ in the same observational class. The choice of the experimental method of estimating the causal class in Step~\ref{alg:cslearn_estimate} is left to the user and depends on the behaving agent and the behavior in question. If, for example, $T$ represents whether the spiking rate of a recorded neuron is above a fixed threshold, estimating $P(T\mid \man{I=i})$ could consist of recording the neuron's response to $i$ in a laboratory setting multiple times, and then calculating the probability of spiking from the finite sample. The causal dataset created in Step~\ref{alg:cslearn_coarsen} consists of the observational inputs and their causal classes. The causal dataset is acquired through $\mathcal{O}(N)$ experiments, where $N$ is the number of observational classes. The final step of the algorithm trains a neural network that predicts the causal labels on unseen images. The choice of the method of training is again left to the user.

\subsection{CAUSAL FEATURE MANIPULATION}
\label{sec:manipulator_practice}
Once we have learned $C$ we can use the causal neural network to create synthetic examples of images as similar as possible to the originals, but with a different causal label. The meaning of ``as similar as possible'' depends on the image metric $d$ (see Definition~\ref{def:manipulator}). The choice of $d$ is task-specific and crucial to the quality of the manipulations. In our experiments, we use a metric induced by an $L_2$ norm. Alternatives include other $L_p$-induced metrics, distances in implicit feature spaces induced by image kernels~\citep{Harchaoui2007,Grauman2007,Bosch2007,Vishwanathan2010} and distances in learned representation spaces~\citep{Bengio2013}.

Algorithm~\ref{alg:manipulator} proposes one way to learn the manipulator function using a simple manipulation procedure that approximates the requirements of Definition~\ref{def:manipulator} up to local minima.
\IncMargin{1.5em}
\begin{algorithm}[h!t]
\caption{Manipulator Function Learning}
\label{alg:manipulator}
\SetKwData{Nq}{Q}\SetKwData{nIters}{nIters}
\SetKwData{iter}{iter}
\SetKwData{Data}{data}\SetKwData{NN}{NN}
\SetKwFunction{Train}{Train}\SetKwFunction{AgentQ}{A}
\SetKwInOut{Input}{input}
\SetKwInOut{Output}{output}

\Input{$d\colon \mathcal{I}\times\mathcal{I}\rightarrow \mathbb{R}_+$ 
           -- a metric on the image space\\
       $\mathcal{D}_{csl} = \{(i_1, c_1),\cdots (i_N, c_N)\}$ 
           -- causal data \\
       $\mathcal{C} = \{C_1, \cdots, C_M\}$ -- the set of causal\\ 
       classes (so that $\forall_i c_i \in \mathcal{C}$)\\
       $\Train$ -- a neural net training algorithm\\
       $\nIters$ -- number of experiment iterations\\
       $\Nq$ -- number of queries per iteration\\
       $\alpha$ -- manipulation tuning parameter\\
       $\AgentQ \colon \mathcal{I}\rightarrow\mathcal{C}$ 
           -- an oracle for $P(T \mid  \Do{I})$}
\Output{$M_C\colon \mathcal{I}\times\mathcal{C}\to \mathcal{I}$ -- the manipulator function}
\BlankLine

\For{$l \leftarrow 1$ \KwTo $ \nIters$}
{
  $C \leftarrow \Train(\mathcal{D}_{csl})$\;\label{alg:manipulator:train_csl}
  Choose manipulation starting points $\phantom{WWW}\{i_{l,1},\cdots,i_{l,\Nq}\}$ at random from $\mathcal{D}_{csl}$\;
  Choose manipulation targets $\{\hat{c}_{l,1}, \cdots, \hat{c}_{l,\Nq}\}$ $\phantom{WWW}$such that $\hat{c}_{l,k}\neq c_{l,k}$\;
  \For{$k\leftarrow 1$ \KwTo $\Nq$}
  {
    $\hat{\imath}_{l,k} \leftarrow \argmin{j\in \mathcal{I}}(1-\alpha)|C(j)-\hat{c}_{l,k}|$ 
    $\phantom{WWWWWWWWW}+\alpha\; d(j, i_{l,k})$\;\label{alg:manipulator:mainLoop}
  }
  $\mathcal{D}_{csl} \leftarrow \mathcal{D}_{csl} \cup \{(\hat{\imath}_{l,1}, \AgentQ(\hat{\imath}_{l,1})), \cdots, $ $\phantom{WWWWWWWWW}(\hat{\imath}_{l,\Nq}, \AgentQ(\hat{\imath}_{l,\Nq}))\}$\;\label{alg:manipulator:end_loop}
}
\end{algorithm}\DecMargin{-1.5em}
The algorithm, inspired by the active learning techniques of uncertainty sampling~\citep{Lewis1994} and density weighing~\citep{Settles2008}, starts off by training a causal neural network in Step~\ref{alg:manipulator:train_csl}. If only observational data is available, this can be achieved using Algorithm~\ref{alg:cslearn}. Next, it randomly chooses a set of images to be manipulated, and their target post-manipulation causal labels. The loop that starts in Step~\ref{alg:manipulator:mainLoop} then takes each of those images and searches for the image that, among the images with the same desired causal class, is closest to the original image. Note that the causal class boundaries are defined by the current causal neural net $C$. Since $C$ is in general a highly nonlinear function and it can be hard to find its inverse sets, we use an approximate solution. The algorithm thus finds the minimum of a weighted sum of $|C(j)-\hat{c}_{l,k}|$ (the difference of the output image $j$'s label and the desired label $\hat{c}_{l,k}$) and $d(i_{l,k}, j)$ (the distance of the output image $j$ from the original image $i_{l,k}$).

At each iteration, the algorithm performs $Q$ manipulations and the same number of causal queries to the agent, which result in new datapoints $(\hat{\imath}_{l,1}, A(\hat{\imath}_{l,1})), \cdots, (\hat{\imath}_{l,Q}, A(\hat{\imath}_{l,Q}))$. It is natural to claim that the manipulator performs well if $A(\hat{\imath}_{l,k}) \approx \hat{c}_{l,k}$ for many $k$, which means the target causal labels agree with the true causal labels. We thus define the \emph{manipulation error} of the $l$th iteration $M\!Err_{l}$ as
\begin{equation}
  M\!Err_{l} = \frac{1}{Q}\sum_{k=1}^{Q} |A(\hat{\imath}_{l,k}) - \hat{c}_{l,k}|.\label{eq:manipulation_error}
\end{equation}
While it is important that our manipulations are accurate, we also want them to be minimal. Another measure of interest is thus the \emph{average manipulation distance} 

\begin{equation}
  M\!Dist_{l} = \frac{1}{Q}\sum_{k=1}^{Q} d(I_{l,k}, \hat{\imath}_{l,k}).\label{eq:manipulation_distance}
\end{equation}

A natural variant of Algorithm~\ref{alg:manipulator} is to set $nIters$ to a large integer and break the loop when one or both of these performance criteria reaches a desired value.

\section{EXPERIMENTS}
\label{sec:experiments} 
In order to illustrate the concepts presented in this article we perform two causal feature learning experiments. The first experiment, called {\sc grating}, uses observational and causal data generated by the model from Section~\ref{sec:example}. The {\sc grating} experiment confirms that our system can learn the ground truth cause and ignore the spurious correlates of a behavior. The second experiment, {\sc mnist}, uses images of hand-written digits~\citep{LeCun1998} to exemplify the use of the manipulator function on slightly more realistic data: in this example, we transform an image into a maximally similar image with another class label. 

We chose problems that are simple from the computer vision point of view. Our goal is to develop the theory of visual causal feature learning and show that it has feasible algorithmic solutions; we are at this point not engineering advanced computer vision systems.

\subsection{THE {\sc GRATING} EXPERIMENT}
\begin{figure}
\centering
\includegraphics{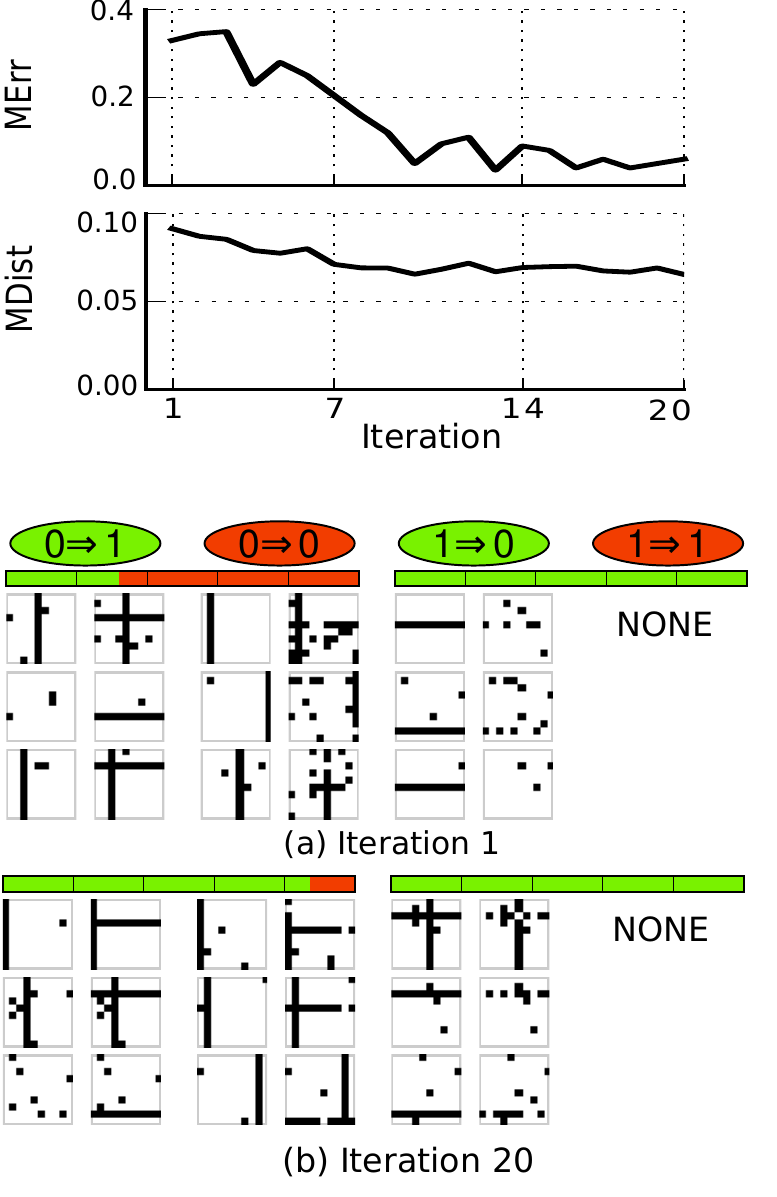}
\captionof{figure}{Manipulator learning for {\sc grating.} \textbf{Top.} The plots show the progress of our manipulator function learning algorithm over ten iterations of experiments for the {\sc grating} problem. The manipulation error decreases quickly with progressing iterations, whereas the manipulation distance stays close to constant. \textbf{Bottom.} Original and manipulated {\sc grating} images. See text for the details.}
\label{fig:grating_results}
\end{figure}

In this experiment we generate data using the model of Fig.~\ref{fig:grating_data}, with two minor differences: $H_1$ and $H_2$ only induce one v-bar or h-bar in the image and we restrict our observational dataset to images with only about 3\% of the pixels filled with random noise (see Fig.~\ref{fig:grating_results}). Both restrictions increase the clarity of presentation. We use Algorithms~\ref{alg:cslearn}~and~\ref{alg:manipulator} (with minor modifications imposed by the binary nature of the images) to learn the visual cause of behavior $T$. 

Figure~\ref{fig:grating_results} (top) shows the progress of the training process. The first step (not shown in the figure) uses the CCT to learn the causal labels on the observational data. We then train a simple neural network (a fully connected network with one hidden layer of 100 units) on this data. The same network is used on Iteration 1 to create new manipulated exemplars. We then follow Algorithm~\ref{alg:manipulator} to train the manipulator iteratively. Fig.~\ref{fig:grating_results} (bottom) illustrates the difference between the manipulator on Iteration 1 (which fails almost 40\% of the time) and Iteration 20, where the error is about 6\%. Each column shows example manipulations of a particular kind. Columns with green labels indicate successful manipulations of which there are two kinds: switching the causal variable on ($0 \Rightarrow 1$, ``adding the h-bar''), or switching it off ($1\Rightarrow 0$, ``removing the h-bar''). Red-labeled columns show cases in which the manipulator failed to influence the cause: That is, each red column shows an original image and its manipulated version which the manipulator believes should cause a change in $T$, but which does not induce such change. The red/green horizontal bars show the percentage of success/error for each manipulation direction. Fig.~\ref{fig:grating_results} (bottom, a) shows that after training on the causally-coarsened observational dataset, the manipulator fails about 40\% of the time. In Fig.~\ref{fig:grating_results} (b), after twenty manipulator learning iterations, only six manipulations out of a hundred are unsuccessful. Furthermore, the causally irrelevant image pixels are also much better preserved than at iteration 1. The fully-trained manipulator correctly learned to manipulate the presence of the h-bar to cause changes in $T$, and ignores the v-bar that is strongly correlated with the behavior but does not cause it.

\subsection{THE {\sc MNIST ON MTURK} EXPERIMENT}
In this experiment we start with the {\sc MNIST} dataset of handwritten digits. In our terminology, this -- as well as any standard vision dataset -- is already causal data: the labels are assigned in an experimental setting, not ``in nature''. 

\begin{figure}
\centering
\includegraphics{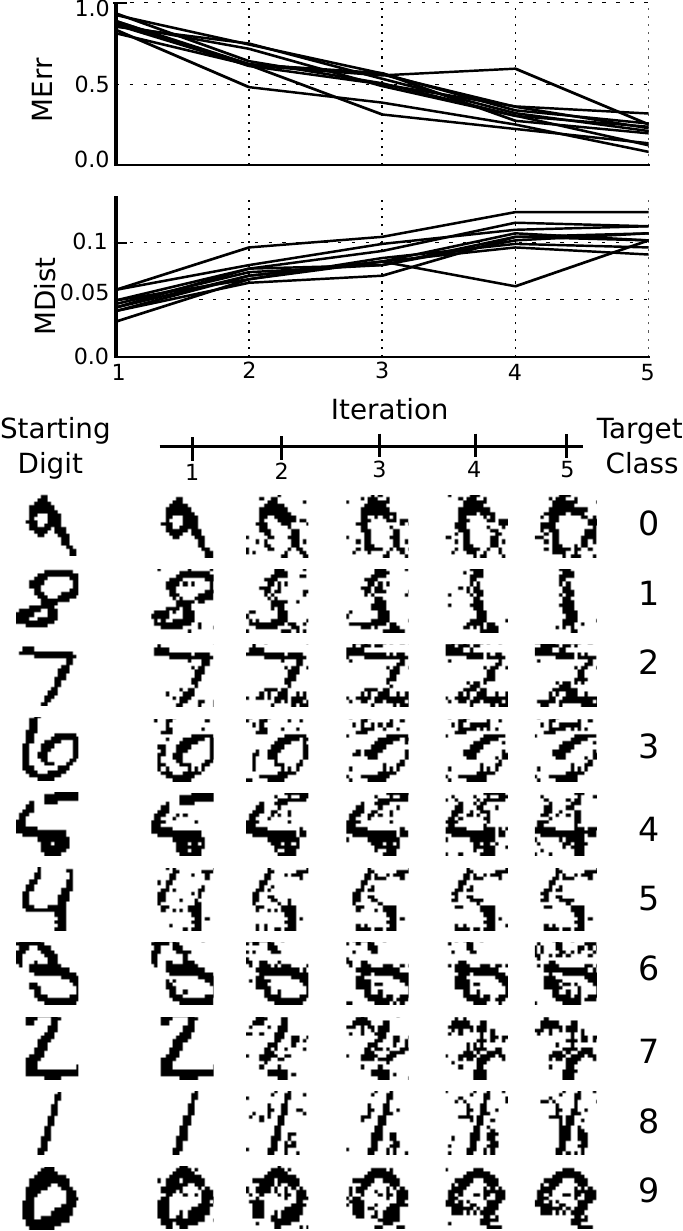}
\caption{Manipulator Learning for {\sc mnist on mturk.} \textbf{Top.} In contrast to the {\sc grating} experiment, here the manipulation distance grows as the manipulation error decreases. This is because a successful manipulator needs to change significant parts of each image (such as continuous strokes). \textbf{Bottom.} Visualization of manipulator training on randomly selected (not cherry-picked) {\sc mnist} digits. See text for the details.}
\label{fig:mnist_results}
\end{figure}

Consider the following binary human behavior: $T = 1$ if a human observer answers affirmatively to the question ``Does this image contain the digit `7'?'', while $T=0$ if the observer judges that the image does not contain the digit `7'. For simplicity we will assume that for any image either $P(T=1 \mid  \man{I})=0$ or $P(T=1 \mid  \man{I})=1$. Our task is to learn the manipulator function that will take any image and modify it minimally such that it will become a `7' if it was not before, or will stop resembling a `7' if it did originally. 

We conduct the manipulator training separately for all the ten {\sc mnist} digits using human annotators on Amazon Mechanical Turk. The exact training procedure is described in Appendix 10. Fig.~\ref{fig:mnist_results} (top) shows training progress. As in Fig.~\ref{fig:grating_results}, the manipulation error decreases with training. Fig.~\ref{fig:mnist_results} (bottom) visualizes the manipulator training progress.  In the first row we see a randomly chosen {\sc MNIST} ``9'' being manipulated to resemble a ``0'', pushed through successive ``0-vs-all'' manipulators trained at iterations 0, 1, ..., 5 (iteration~1 shows what the neural net takes to be the closest manipulation to change the ``9'' to a ``0'' purely on the basis of the non-manipulated data). Further rows perform similar experiments for the other digits. The plots show how successive manipulators progressively remove the original digits' features and add target class features to the image.
\vspace{-0.1in}
\section{DISCUSSION}

\label{sec:discussion}
We provide a link between causal reasoning and neural network models that have recently enjoyed tremendous success in the fields of machine learning and computer vision~\citep{LeCun1998, Russakovsky2014}. Despite very encouraging results in image classification~\citep{Krizhevsky2012}, object detection~\citep{Dollar2012} and fine-grained classification~\citep{Branson2014,Zhang2014}, some researchers have found that visual neural networks can be easily fooled using adversarial examples \citep{Fergus2014,Goodfellow2014}. The learning procedure for our manipulator function could be viewed as an attempt to train a classifier that is robust against such examples. The procedure uses causal reasoning to improve on the boundaries of a standard, correlational classifier (Fig.~\ref{fig:grating_results}~and~\ref{fig:mnist_results} show the improvement). However, the ultimate purpose of a causal manipulator network is to extract truly causal features from data and automatically perform causal manipulations based on those features.

A second contribution concerns the field of causal discovery. Modern causal discovery algorithms presuppose that the set of causal variables is well-defined and meaningful. What exactly this presupposition entails is unclear, but there are clear counter-examples: $x$ and $2x$ cannot be two distinct causal variables. There are also well understood problems when causal variables are aggregates of other variables~\citep{Chu2003,Spirtes2004a}. We provide an account of how causal macro-variables can supervene on micro-variables. 

This article is an attempt to clarify how one may construct a set of well-defined causal macro-variables that function as basic relata in a causal graphical model. This step strikes us as essential if causal methodology is to be successful in areas where we do not have clearly delineated candidate causes or where causes supervene on micro-variables, such as in climate science and neuroscience, economics and---in our specific case---vision. 
\vspace{-0.1cm}
\subsubsection*{Acknowledgements}
KC's work was funded by the Qualcomm Innovation Fellowship 2014. KC's and PP's work was supported by the ONR MURI grant N00014-10-1-0933. FE would like to thank Cosma Shalizi for pointers to many relevant results this paper builds on. 

\section{APPENDIX: PROOF OF THE CAUSAL COARSENING THEOREM}
Before we prove the Causal Coarsening Theorem, we prove its less general version in order to split the rather complex proof of CCT into two parts. This Auxiliary Theorem can be proven using simpler techniques, however here we deliberately use techniques that transfer directly to the proof of the CCT.

\vspace{1em}
\noindent{\bf Auxiliary Theorem} {\it Among all the generative models of the form discussed in Fig. 2 (in the main text), the subset of distributions $P(T, H, I)$ for which the causal partition is not a coarsening (proper or improper) of the observational partition is Lebesgue measure zero.}
\vspace{0.5em}

\begin{proof}\label{thm:cct1}
Our proof is inspired by a proof used by \citet{Meek1995} to prove that almost all distributions compatible with a given causal graph are faithful. The proof strategy is thus first to express the proposition that for a given distribution, the observational partition does not refine the causal partition as a polynomial equation on the space of all distributions compatible with the model. We then show that this polynomial equation is not trivial, i.e.\ there is at least one distribution that is not its root. By a simple algebraic lemma, this will prove the theorem. We extend \citeauthor{Meek1995}'s proof technique in our usage of Fubini's Theorem for the Lebesgue integral. It allows us to ``split'' the polynomial constraint into multiple different constraints along several of the distribution parameters. This allows for additional flexibility in creating useful assumptions (in our proof, the assumption that the datapoints have well-defined causal classes, but the observational class can still vary freely). 

Assume that $T$ is binary and $H=(H_1,\cdots,H_M)$, $I$ are discrete variables (say $|H_i|=K_i, |I|=N$, though $N$ can be very large. We will use the notation $K \triangleq K_1\times\cdots\times K_M$ for simplicity later on). The discreteness assumption is not crucial, but will simplify the reasoning. We can factorize the joint as $P(T,H,I)=P(T\mid H,I)P(I\mid H)P(H)$. $P(T\mid H,I)$ can be parametrized by $ |H_1|\times\cdots\times |H_M|\times |I|=K\times N$ parameters, $P(I\mid H)$ by $(N-1)\times K$ parameters, and $P(H)$ by another $K$ parameters, all of which are independent.
Call the parameters, respectively, 
\begin{align*}
\alpha_{h,i} &\triangleq P(T=0 \mid H=h, I=i)\\
\beta_{i,h} &\triangleq P(I=i\mid H=h)\\
\gamma_h &\triangleq P(H=h)
\end{align*}
We will denote parameter vectors as
\begin{align*}
\alpha &= (\alpha_{h_1,i_1}, \cdots, \alpha_{h_K,i_N}) \in\mathbb{R}^{K\times N} \\
\beta &= (\beta_{i_1,h_1}, \cdots, \beta_{i_{N-1}, h_K}) \in\mathbb{R}^{(N-1)\times K}\\
\gamma &= (\gamma_{h_1}, \cdots, \gamma_{h_K}) \in\mathbb{R}^{K},
\end{align*}
where the indices are arranged in lexicographical order. This creates a one-to-one correspondence of each possible joint distribution $P(T,H,I)$ with a point $(\alpha,\beta,\gamma)\in P[\alpha,\beta,\gamma] \subset \mathbb{R}^{K^3\times N\times (N-1)}$, where $P[\alpha, \beta, \gamma]$ is the $K^3\times N\times (N-1)$-dimensional simplex of multinomial distributions.

To proceed with the proof, we first pick any point in the $P(T\mid H,I) \times P(H)$ space: that is, we fix the values of $\alpha$ and $\gamma$. The only free parameters are now $\beta_{i,h}$ for all values of $i,h$; varying these values creates a subset of the space of all the distributions which we will call 
\begin{equation*}
P[\beta; \alpha,\gamma]=\{(\alpha, \beta,\gamma)\; \mid \; \beta \in [0,1]^{(N-1)\times K}\}.
\end{equation*}
$P[\beta; \alpha, \gamma]$ is a subset of $P[\alpha,\beta,\gamma]$ isometric to the $[0,1]^{(N-1)\times K}$-dimensional simplex of multinomials. We will use the term $P[\beta; \alpha, \gamma]$ to refer both the subset of $P[\alpha,\beta,\gamma]$ and the lower-dimensional simplex it is isometric to, remembering that the latter comes equipped with the Lebesgue measure on $\mathbb{R}^{(N-1)\times K}$.

Now we are ready to show that the subset of $P[\beta; \alpha, \gamma]$ which does not satisfy the Causal Coarsening constraint is of measure zero with respect to the Lebesgue measure. To see this, first note that since $\alpha$ and $\gamma$ are fixed, each image $i$ has a well-defined causal class $C(i) = \sum_h \alpha_{h,i}\gamma_{h}$. The Causal Coarsening constraint says ``For every pair of images $i,j$ such that $P(T \mid  i) = P(T \mid  j)$ it holds that $C(i) = C(j)$.'' The subset of $P[\beta;\alpha, \gamma]$  of all distributions that do not satisfy the constraint consists of the $P(T, H, I)$ for which for some $i,j$ it holds that 
\begin{align*}
P(T=0\mid i)=P(T=0\mid j) \quad & \textrm{and} \quad C(i)\neq C(j).
\end{align*}
Take any pair $i,j$ for which $C(i)\neq C(j)$ (if such a pair does not exist, then the Causal Coarsening constraint holds for all the distributions in $P[\beta;\alpha, \gamma]$). We can write 

\begin{align*}
P(T=0 \mid  i) &= \sum_h P(T=0\mid h, i) P(h\mid i)\\
         &= \frac{1}{P(i)}\sum_h P(T=0 \mid  h,i)P(i\mid h)P(h).
\end{align*}

Since the same equation applies to $P(T=0 \mid  j)$, the constraint $P(T\mid i) = P(T\mid j)$ can be rewritten

\begin{align*}
\frac{1}{P(i)}&\sum_h P(T=0 \mid  h,i)P(i\mid h)P(h)\\
  &= \frac{1}{P(j)}\sum_h P(T=0 \mid  h,j)P(j\mid h)P(h)\\
\iff &P(j)\sum_h P(T=0 \mid  h,i)P(i\mid h)P(h)\\
    & - P(i)\sum_h P(T=0 \mid  h,j)P(j\mid h)P(h) =0,
\end{align*}
which we can rewrite in terms of the independent parameters (after defining $\alpha_{0,h,i}=\alpha_{h,i}$ and $\alpha_{1,h,i}=1-\alpha_{h,i}$) and further simplify as

\begin{align*}
\left(\sum_{t\in\{0,1\}}\sum_h\alpha_{t,h,j}\gamma_{h}\beta_{j,h}\right)\sum_h\alpha_{0,h,i}\gamma_{h}\beta_{i,h}-\phantom{.WWWW}\notag\\
-\left(\sum_{t\in\{0,1\}}\sum_h\alpha_{t,h,i}\gamma_{h}\beta_{i,h}\right)\sum_h\alpha_{0, h,j}\gamma_{h}\beta_{j,h} = 0\notag\\
\iff\left(\sum_h\alpha_{1,h,j}\gamma_{h}\beta_{j,h}\right)\sum_h\alpha_{0,h,i}\gamma_{h}\beta_{i,h}-\phantom{WWWWW}\notag\\
 -\left(\sum_h\alpha_{1,h,i}\gamma_{h}\beta_{i,h}\right)\sum_h\alpha_{0, h,j}\gamma_{h}\beta_{j,h} = 0\notag\\
\end{align*}
\begin{align}
\iff &\left(\sum_h(1-\alpha_{h,j})\gamma_{h}\beta_{j,h}\right)\sum_h\alpha_{h,i}\gamma_{h}\beta_{i,h}-\phantom{WWWW}\notag\\
&-\left(\sum_h(1-\alpha_{h,i})\gamma_{h}\beta_{i,h}\right)\sum_h\alpha_{h,j}\gamma_{h}\beta_{j,h} = 0\notag\\
\iff&\left(\sum_h\gamma_{h}\beta_{j,h}\right)\sum_h\alpha_{h,i}\gamma_{h}\beta_{i,h}-\phantom{WWWWW}\notag\\
&-\left(\sum_h\gamma_{h}\beta_{i,h}\right)\sum_h\alpha_{h,j}\gamma_{h}\beta_{j,h} = 0,\label{eq:constraint}
\end{align}

which is a polynomial constraint on $P[\beta;\alpha, \gamma]$ (note that to keep the notation manageable, we have omitted the dependent term $1-\sum_h\gamma_h$ from the equations). By a simple algebraic lemma \citep[proven by][]{Okamoto1973}, if the above constraint is not trivial (that is, if there exists $\beta$ for which the constraint does not hold), the subset of $P[\beta;\alpha, \gamma]$ on which it holds is measure zero.

To see that Eq.~\eqref{eq:constraint} does not always hold, note that if for \emph{any} $h^*$ we set $\beta_{i,h^*}=1$ (and thus $\beta_{i,h}=0$ for any $h\neq h^*$) and $\beta_{j,h^*}=1$, the equation reduces to 
\begin{align*}
(\gamma_{h^*})^2(\alpha_{h_i,i}-\alpha_{h_j,h}) = 0.
\end{align*}
Thus if Eq.~\eqref{eq:constraint} was trivially true, we would have $\alpha_{h,i}=\alpha_{h,j}$ or $\gamma_h=0$ for all $h$. However, this implies $C(i) = C(j)$, which contradicts our assumption.

We have now shown that the subset of $P[\beta;\alpha, \gamma]$ which consists of distributions for which $P(T\mid i)=P(T\mid j)$ (even though $C(i) \neq C(j)$) is Lebesgue measure zero. Since there are only finitely many pairs of images $i,j$ for which $C(i)\neq C(j)$, the subset of $P[\beta;\alpha, \gamma]$ of distributions which violate the Causal Coarsening constraint is also Lebesgue measure zero. The remainder of the proof is a direct application of Fubini's theorem. 

For each $\alpha, \gamma$, call the (measure zero) subset of $P[\beta;\alpha, \gamma]$ that violates the Causal Coarsening constraint $z[\alpha, \gamma]$. Let $Z=\cup_{\alpha, \gamma} z[\alpha, \gamma] \subset P[\alpha, \beta, \gamma]$ be the set of all the joint distributions which violate the Causal Coarsening constraint. We want to prove that $\mu(Z)=0$, where $\mu$ is the Lebesgue measure. To show this, we will use the indicator function 

\begin{equation*}
\hat{z}(\alpha, \beta, \gamma) = \left\{
  \begin{array}{l} 1\quad\text{if } \beta \in z[\alpha,\gamma],\\
                    0\quad\text{otherwise}.
  \end{array}
  \right.
\end{equation*}

By the basic properties of positive measures we have 

\begin{equation*}
\mu(Z) = \int_{P[\alpha, \beta, \gamma]}\hat{z}\;d\mu.
\end{equation*}

It is a standard application of Fubini's Theorem for the Lebesgue integral to show that the integral in question equals zero. For simplicity of notation, let
\begin{align*}
\mathcal{A} &= \mathbb{R}^{K\times N} \\
\mathcal{B} &= \mathbb{R}^{N\times K}\\
\mathcal{G} &= \mathbb{R}^{K}.
\end{align*}

We have 

\begin{align}
\int_{P[\alpha, \beta, \gamma]}\hat{z}\;d\mu 
  &= \int_{\mathcal{A}\times\mathcal{B}\times\mathcal{G}} \hat{z}(\alpha, \beta, \gamma)\, d(\alpha, \beta, \gamma)\notag\\
  &= \int_{\mathcal{A}\times\mathcal{G}}\; \int_\mathcal{B}\hat{z}(\alpha, \beta, \gamma)\, d(\beta)\; d(\alpha, \gamma)\notag \\
  &= \int_{\mathcal{A}\times\mathcal{G}}\; \mu(z[\alpha, \gamma])\; d(\alpha, \gamma) \label{eq:measureZero:indicator}\\
  &= \int_{\mathcal{A}\times\mathcal{G}} 0\, d(\alpha, \gamma)\notag\\
  &= 0\notag.
\end{align}
Equation~\eqref{eq:measureZero:indicator} follows as $\hat{z}$ restricted to $P[\beta;\alpha, \gamma]$ is the indicator function of $z[\alpha, \gamma]$. 

This completes the proof that $Z$, the set of joint distributions over $T, H$ and $I$ that violate the Causal Coarsening constraint, is measure zero.
\end{proof}

We are now ready to prove the main theorem.\\
\\
\noindent{\bf Theorem (Causal Coarsening Theorem)} {\it 
Among all the generative models  of the form discussed in Fig.~2 (in the main text) that have distributions $P(T,\mathbf{H},I)$ that induce some given observational partition $\Pi_o$, \emph{almost all} induce a causal partition $\Pi_c$ that is a coarsening of $\Pi_o$.}
\vspace{0.5em}

\begin{proof}
Any variables that appear in this proof without definition are defined in the proof of the Auxiliary Theorem. We take the same $\alpha, \beta, \gamma$ parametrization of distributions. Fixing an observational partition means fixing a set of observational constraints (OCs)

\begin{align*}
P(T \mid  i^1_1)&=\cdots=P(T\mid i^1_{N_1}),\\
&\vdots\\
P(T \mid  i^L_1)&=\cdots=P(T\mid i^L_{N_K}),
\end{align*}

where $1\leq L\leq N$ is the number of observational classes. Since $P(T,H,I) = P(H \mid  T,I)P(T\mid I)P(I)$, $P(T\mid i)$ is an independent parameter in the unrestricted $P(T,H,I)$, and the OCs reduce the number of independent parameters of the joint by $\sum_{l=1}^L (N_l-1)$. We want to express this parameter-space reduction in terms of the $\alpha,\beta$ and $\gamma$ parameterization and then apply the proof of the Auxiliary Theorem. To do this, for each observational class $l$, choose a representative image $\hat{\imath}^l$ such that 
\begin{align*}
P(T \mid  i^l_m) = P(T \mid  \hat{\imath}^l) \quad \forall_{m\in 1\cdots N_k}.
\end{align*} 
Then for each $i^l_m \neq \hat{\imath}^l$ it holds that

\begin{equation*}
P(T,i^l_m) = P(T\mid \hat{\imath}^l)P(i^l_m)
\end{equation*}
or
\begin{equation*}
\sum_h P(T,h,i^l_m) = P(T\mid \hat{\imath}^l)\sum_h P(h, i^l_m).
\end{equation*}
Picking an arbitrary $h_0$, we can separate the left-hand side as
\begin{equation*}
P(T,h_0,i^l_m) = P(T\mid \hat{\imath}^l)\sum_h P(h, i^l_m)-\sum_{h\neq h_0} P(T,h,i^l_m).
\end{equation*}

Finally, this equation can be rewritten in terms of $\alpha,\beta$ and $\gamma$ as
\begin{equation}
\alpha_{h_0,i}\beta_{i,h_0}\gamma_{h_0} = P(T\mid \hat{\imath}^l)\sum_h \beta_{h,i^l_m}\gamma_h-\sum_{h\neq h_0} \alpha_{h,i^l_m}\beta_{i^l_m}\gamma_h\notag,
\end{equation}
or
\begin{equation*}
\alpha_{h_0,i} = \frac{\left(P(T\mid \hat{\imath}^l)\sum_h \beta_{h,i^l_m}\gamma_h-\sum_{h\neq h_0} \alpha_{h,i^l_m}\beta_{i^l_m}\gamma_h\right)}{\beta_{i,h_0}\gamma_{h_0}}\notag
\end{equation*}

for any $i^l_m \neq \hat{\imath}^l$. There are precisely $\sum_{l=1}^L (N_l-1)$ such equations, altogether equivalent to the observational constraints. Thus we can express any $P(T,H,I)$ distribution that is consistent with a given observational partition in terms of the full range of $\beta$ and $\gamma$ parameters, and a restricted number of independent $\alpha$ parameters. The rest of the proof now follows similarily to the proof of the Auxiliary Theorem and shows that within this restricted parameter space, the parameters for which the (fixed) observational partition is not a refinement of the causal partition is measure zero.
\end{proof}

\section{APPENDIX: CCT EXAMPLES AND COUNTER-EXAMPLES}
\begin{figure}
\includegraphics[width=0.5\textwidth]{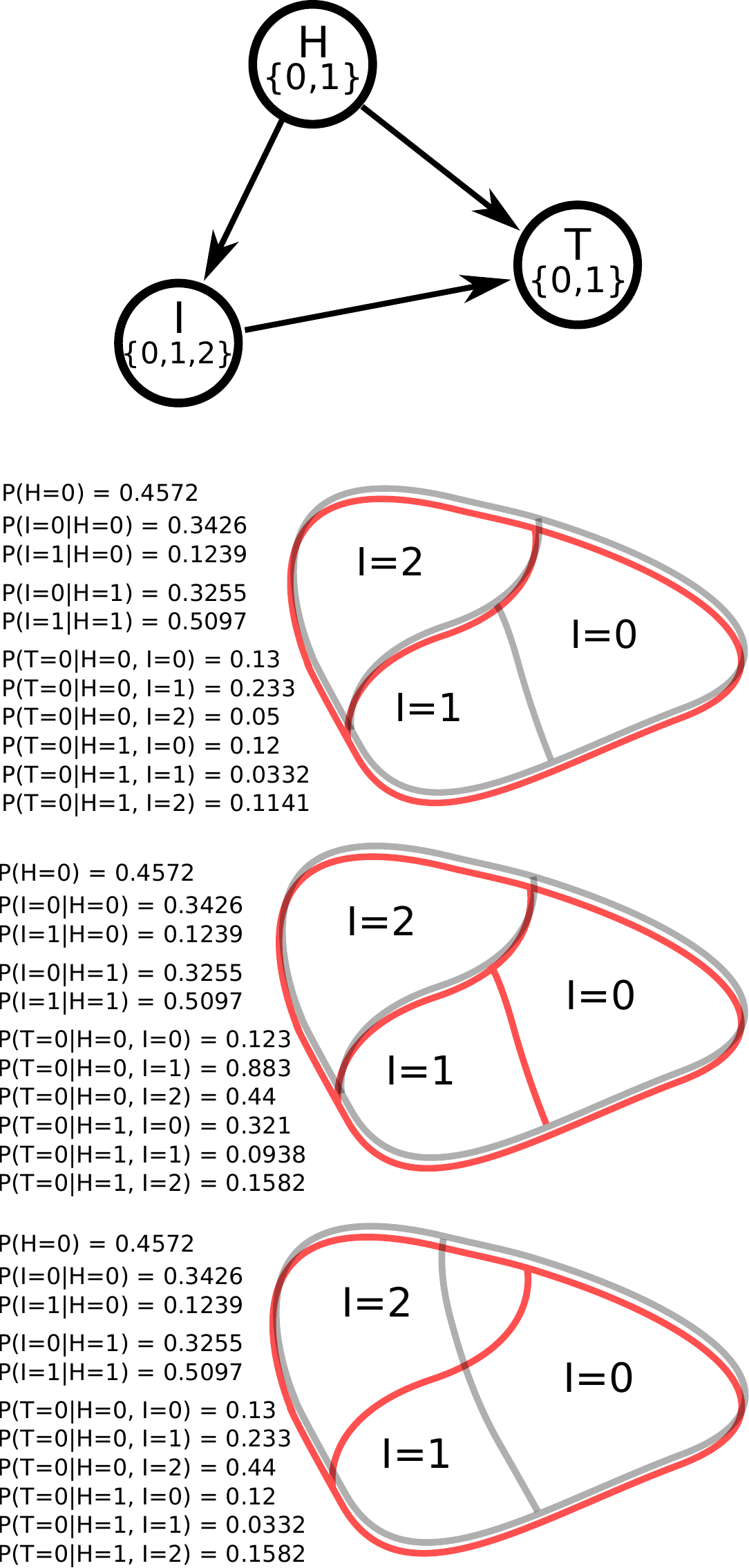}
\caption{A graphical causal model and three faithful probability tables. The first (from the top) table induces a causal partition (red) that is a coarsening of the observational partition (gray) -- specifically, as the figure shows, $P(T | I=0)\neq P(T | I=1)$ but $P(T|\man{I=0}) = P(T|\man{I=1})$. The second table induces an observational partition that is a corasening of the causal partition. The last table induces a causal and an observational partition such that neither is a coarsening of the other. }
\label{fig:counter_examples}
\end{figure}

In Fig.~\ref{fig:counter_examples} we provide examples of three distributions over binary variables $H, T$ and three-valued $I$. The first model induces a causal partition that is a proper coarsening of the observational partition, and thus agrees with the CCT. The second model induces an observational partition that is a proper coarsening of the causal partition -- CCT implies that this is a measure-zero case and that, after fixing the observational partition, we had to carefully tweak the parameters to align the causal partition as it is. The third model induces causal and observational partitions that are incompatible -- that is, neither is a coarsening of the other. This is also a measure-zero case. We provide a Tetrad (\url{http://www.phil.cmu.edu/tetrad/}) file that contains these three models at  \url{http://vision.caltech.edu/~kchalupk/code.html}. It can be used to verify our observational and causal partition computations.

\section{APPENDIX: PROOF OF THE COMPLETE MACRO-VARIABLE DESCRIPTION THEOREM}
\noindent{\bf Theorem (Complete Macro-variable Description)}
{\it
The following two statements hold for $C$ and $S$ as defined in the main text:
\begin{enumerate}
\item $P(T\mid I) = P(T\mid C,S).$
\item Any other variable $X$ such that $P(T\mid I)=P(T\mid X)$ has Shannon entropy $H(X) \geq H(C,S)$.
\end{enumerate}
}
\vspace{0.5em}

\begin{proof}
The first part follows by construction of $S$. For the second part, note that by the CCT there is a bijective correspondence between the pairs of values $(c, s)$ and the observational probabilities $P(T \mid I)$. Call this correspondence $f$, that is $f(c,s) = P(T\mid c,s)$ and $f^{-1}(p) = (c, s\;\text{ s.t. }P(T|c,s)=p)$. Further, define $g$ as the function on $X,$ with $g\colon x\mapsto P(T\mid x)$. But since $P(T\mid X) = P(T\mid I)$, we have $(c,s)=f^{-1}(g(x))$. That is, the value of $C$ and $S$ is a function of the value of $X$, and thus the entropy of $C$ and $S$ is smaller than the entropy of $X$. 
\end{proof}

\section{APPENDIX: PREDICTIVE NON-CAUSAL INFORMATION IN CAUSAL VARIABLE $C$}
In some cases $C$ retains predictive information that is not causal. Consider the following example: We have a causal graph consisting of three variables $\{I,T,H\}$ where the causal relations are $I \rightarrow T$ and $I \leftarrow H\rightarrow T$. All three variables are binary and we have a positive distribution over the variables. In the general case, distributions over this graph satisfy
\begin{enumerate}
\item $P(T|do(I=1)) \neq P(T|do(I=0))$
\item $P(T|I=1) \neq P(T|I=0)$ , and importantly
\item $P(T|I) \neq P(T|do(I))$. 
\end{enumerate}
If we view $I$ as an image (which can either be all black or all white), $T$ as the target behavior and $H$ as a hidden confounder, analogous to the set-up in the main article, then the observational partition $\Pi_o$ has just two classes, namely $\{1,0\}$. But in this case the observational partition \emph{is the same} as the causal partition:  $\Pi_o = \Pi_c$. So by our definition of a spurious correlate, $S$ is a constant, since there are no further distinctions to be made within any of the causal classes. $S$ would be omitted from any standard causal model. Nevertheless, we have in our model still that $P(T|C) \neq P(T|do(C))$, i.e.\ the causal variable $C$ still contains predictive information that is not causal. Given that there is by construction no other than the causal and the trivial partition in this example, it must be the case that $C$ retains predictive non-causal information. It follows that in our definitions of $C$ and $S$, it is not the case that the predictive non-causal components of an image can always be completely separated from the causal features.

\vskip 0.2in

\section{APPENDIX:THE {\sc mnist on mturk} EXPERIMENT}
For this experiment, we started off by training ten one-vs-all neural nets. We used cross-validation to choose among the following architectures: 100 hidden units (h.u.), 300 h.u. (one layer), 100-100 h.u (two layers), 300-300 h.u. (two layers). We used maxout~\citep{Goodfellow2013a} activations (each of which computed the max of 5 linear functions). For training we used stochastic gradient descent in batches of 50 with 50\% dropout~\citep{Hinton2012} on the hidden units, momentum adjustment from 0.5 to 0.99 at iteration 100, learning rate decaying from 0.1 to 0.0001 with exponential coefficient of 1/0.9998, no weight decay, and we enforced the maximum norm of a column of hidden units to 5. The training stopped after 1000 iterations and the iteration with best validation error was chosen. We used the Pylearn2 package~\citep{Goodfellow2013a} to train the networks. 

This initial training was done on 5000 training points and 1250 validation points (both of which come from the {\sc mnist} dataset) for each machine. The training points were chosen at random to include 2500 images of a specific digit class (that is, 2500 zeros for the first machine, 2500 ones for the second machine and so on), and 2500 images of random other digits for each machine. The validation sets were composed similarly. Each machine then used Algorithm 2 to transform 1000 images of digits \textit{from its training set} into maximally similar images of the opposing class. 
\begin{figure}
\centering
\includegraphics[width=.5\textwidth]{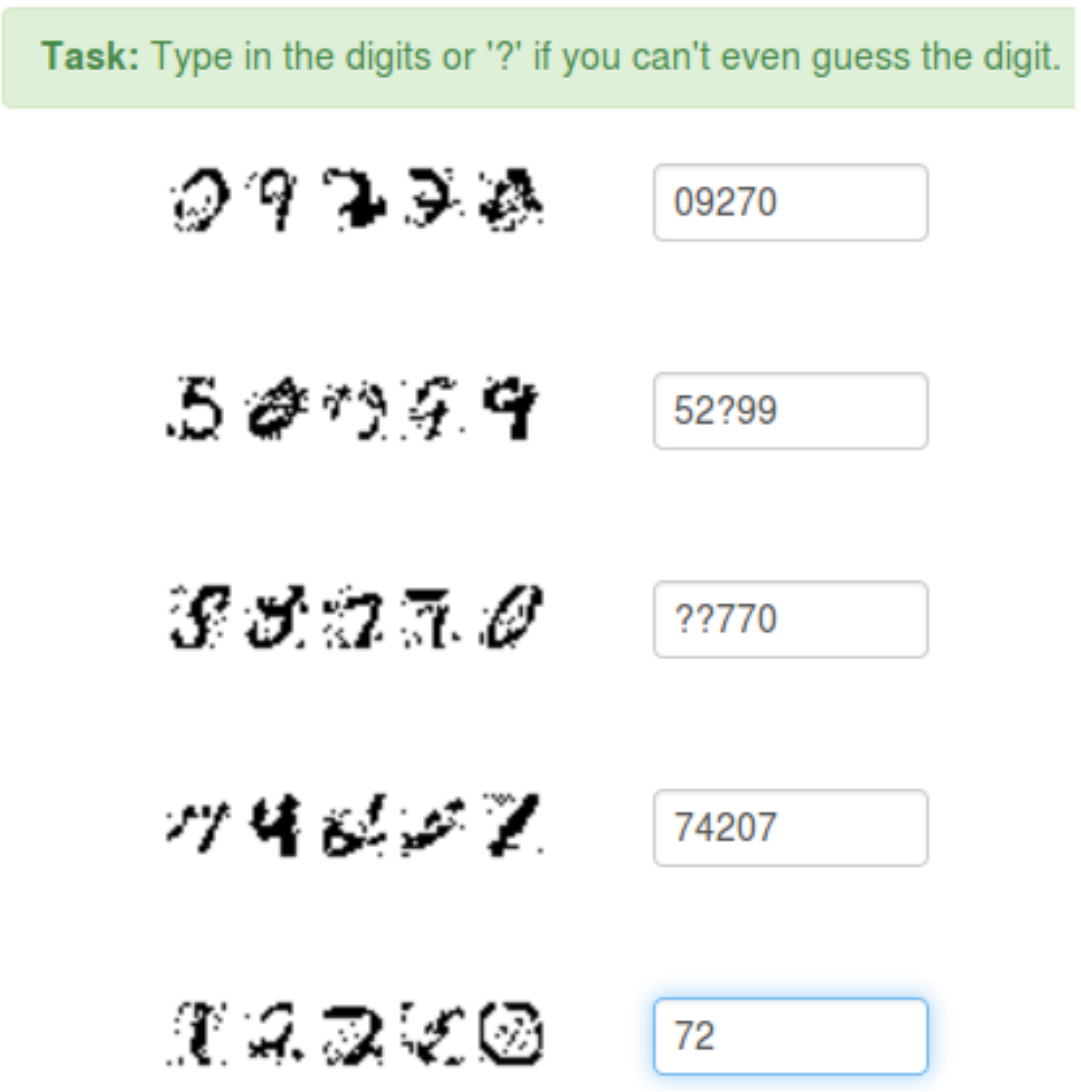}
\caption{The Amazon Mechanical Turk interface we used to query online annotators. An annotator is shown five rows of five manipulated digit images, and is requested to type the digit labels (or `?') into the input boxes. Each annotator goes through ten similar screens, annotating a total of 250 digits.}
\label{fig:MTurkAnnotation}
\end{figure}

We thus started off with ten manipulated datasets of 1000 images each. The first dataset contained images of zeros manipulated to be non-zeros, and all the other digits manipulated to be zeros. The tenth dataset contained images of nines manipulated to be non-nines and the other digits manipulated to be nines. We then used Amazon Mechanical Turk to present all those images to human annotators, using the interface shown in Fig.~\ref{fig:MTurkAnnotation}. The images created by all the manipulator networks were mixed at random together, so that each single annotator (annotating 250 images in one task) would see some images created by each machine. Finally, each of the 10000 images was shown to five annotators; we used 5$\times$40=200 annotators total on each iteration. The annotators labeled the images as either one of the ten digits, or the question mark `?' if there was no recognizable digit in an image. The final label (``target digit'' or ``not target digit'') was chosen using majority of the annotators' votes.

The annotated manipulated digits were then added to the datasets which their respective original images belonged to. We then proceeded to train the next iteration of neural network manipulators on the updated datasets, and so on until completion of the manipulator training.

\subsubsection*{References}
\renewcommand{\section}[2]{}%
\bibliographystyle{plainnat}
\bibliography{bibliography}

\begin{thebibliography}{31}
\providecommand{\natexlab}[1]{#1}
\providecommand{\url}[1]{\texttt{#1}}
\expandafter\ifx\csname urlstyle\endcsname\relax
  \providecommand{\doi}[1]{doi: #1}\else
  \providecommand{\doi}{doi: \begingroup \urlstyle{rm}\Url}\fi

\bibitem[Bengio et~al.(2013)Bengio, Courville, and Vincent]{Bengio2013}
Y.~Bengio, A.~Courville, and P.~Vincent.
\newblock {Representation learning: A review and new perspectives}.
\newblock \emph{Pattern Analysis and Machine Intelligence}, 35\penalty0
  (8):\penalty0 1798--1828, 2013.

\bibitem[Bosch et~al.(2007)Bosch, Zisserman, and Munoz]{Bosch2007}
A.~Bosch, A.~Zisserman, and X.~Munoz.
\newblock {Representing shape with a spatial pyramid kernel}.
\newblock In \emph{6th ACM International Conference on Image and Video
  Retrieval}, pages 401--408, 2007.

\bibitem[Branson et~al.(2014)Branson, Horn, and Wah]{Branson2014}
S.~Branson, G.~Van Horn, and C.~Wah.
\newblock {The Ignorant Led by the Blind: A Hybrid Human–Machine Vision
  System for Fine-Grained Categorization}.
\newblock \emph{International Journal of Computer Vision}, 108\penalty0
  (1-2):\penalty0 3--29, 2014.

\bibitem[Chu et~al.(2003)Chu, Glymour, Scheines, and Spirtes]{Chu2003}
T.~Chu, C.~Glymour, R.~Scheines, and P.~Spirtes.
\newblock {A statistical problem for inference to regulatory structure from
  associations of gene expression measurements with microarrays}.
\newblock \emph{Bioinformatics}, 19\penalty0 (9):\penalty0 1147--1152, 2003.

\bibitem[Dollar et~al.(2012)Dollar, Wojek, Schiele, and Perona]{Dollar2012}
P.~Dollar, C.~Wojek, B.~Schiele, and P.~Perona.
\newblock {Pedestrian detection: An evaluation of the state of the art}.
\newblock \emph{IEEE Transactions on Pattern Analysis and Machine
  Intelligence}, 34\penalty0 (4):\penalty0 743--761, 2012.

\bibitem[Fire and Zhu(2013{\natexlab{a}})]{Fire2013a}
A.~S. Fire and S.~C. Zhu.
\newblock {Using causal induction in humans to learn and infer causality from
  video}.
\newblock \emph{The Annual Meeting of the Cognitive Science Society (CogSci)},
  2013{\natexlab{a}}.

\bibitem[Fire and Zhu(2013{\natexlab{b}})]{Fire2013b}
A.~S. Fire and S.~C. Zhu.
\newblock {Learning Perceptual Causality from Video.}
\newblock \emph{AAAI Workshop: Learning Rich Representations from Low-Level
  Sensors}, 2013{\natexlab{b}}.

\bibitem[Goodfellow and Warde-Farley(2013)]{Goodfellow2013a}
I.~J. Goodfellow and D.~Warde-Farley.
\newblock {Maxout networks}.
\newblock \emph{arXiv preprint arXiv:1302.4389}, 2013.

\bibitem[Goodfellow et~al.(2014)Goodfellow, Shlens, and
  Szegedy]{Goodfellow2014}
I.~J. Goodfellow, J.~Shlens, and C.~Szegedy.
\newblock {Explaining and Harnessing Adversarial Examples}.
\newblock \emph{arXiv preprint arXiv:1412.6572}, 2014.

\bibitem[Grammer and Thornhill(1994)]{Grammer1994}
K.~Grammer and R.~Thornhill.
\newblock {Human (Homo sapiens) facial attractiveness and sexual selection: The
  role of symmetry and averageness.}
\newblock \emph{Journal of Comparative Psychology}, 108\penalty0 (3):\penalty0
  233--242, 1994.

\bibitem[Grauman and Darrell(2007)]{Grauman2007}
K.~Grauman and T.~Darrell.
\newblock {The pyramid match kernel: Efficient learning with sets of features}.
\newblock \emph{Journal of Machine Learning Research}, 8:\penalty0 725--260,
  2007.

\bibitem[Guyon et~al.(2007)Guyon, Elisseeff, and Aliferis]{Guyon2007a}
I.~Guyon, A.~Elisseeff, and C.~Aliferis.
\newblock {Causal feature selection}.
\newblock In \emph{Computational Methods of Feature Selection Data Mining and
  Knowledge Discovery Series}, pages 63--85. Chapman and Hall/CRC, 2007.

\bibitem[Harchaoui and Bach(2007)]{Harchaoui2007}
Z.~Harchaoui and F.~Bach.
\newblock {Image classification with segmentation graph kernels}.
\newblock In \emph{IEEE Computer Society Conference on Computer Vision and
  Pattern Recognition}, pages 1--8, 2007.

\bibitem[Hinton and Srivastava(2012)]{Hinton2012}
G.~E. Hinton and N.~Srivastava.
\newblock {Improving neural networks by preventing co-adaptation of feature
  detectors}.
\newblock \emph{arXiv preprint arXiv:1207.0580}, 2012.

\bibitem[Hoel et~al.(2013)Hoel, Albantakis, and Tononi]{Hoel2013}
E.~P. Hoel, L.~Albantakis, and G.~Tononi.
\newblock {Quantifying causal emergence shows that macro can beat micro}.
\newblock \emph{Proceedings of the National Academy of Sciences}, 110\penalty0
  (49):\penalty0 19790--19795, 2013.

\bibitem[Krizhevsky et~al.(2012)Krizhevsky, Sutskever, and
  Hinton]{Krizhevsky2012}
A.~Krizhevsky, I.~Sutskever, and G.~E. Hinton.
\newblock {ImageNet Classification with Deep Convolutional Neural Networks}.
\newblock In F.~Pereira, C.~J.~C. Burges, L.~Bottou, and K.~Q. Weinberger,
  editors, \emph{Advances in Neural Information Processing Systems 25}, pages
  1097--1105. 2012.

\bibitem[LeCun et~al.(1998)LeCun, Bottou, Bengio, and Haffner]{LeCun1998}
Y.~LeCun, L.~Bottou, Y.~Bengio, and P.~Haffner.
\newblock {Gradient-based learning applied to document recognition}.
\newblock \emph{Proceedings of the IEEE}, 86\penalty0 (11):\penalty0
  2278--2324, 1998.

\bibitem[Lewis and Gale(1994)]{Lewis1994}
D.~D. Lewis and W.~A. Gale.
\newblock {A sequential algorithm for training text classifiers}.
\newblock In \emph{ACM SIGIR Seventeenth Conference on Research and Development
  in Information Retrieval}, pages 3--12, 1994.

\bibitem[Meek(1995)]{Meek1995}
C.~Meek.
\newblock {Strong completeness and faithfulness in Bayesian networks}.
\newblock In \emph{Eleventh Conference on Uncertainty in Artificial
  Intelligence}, pages 411--418, 1995.

\bibitem[Okamoto(1973)]{Okamoto1973}
M.~Okamoto.
\newblock {Distinctness of the eigenvalues of a quadratic form in a
  multivariate sample}.
\newblock \emph{The Annals of Statistics}, 1\penalty0 (4):\penalty0 763--765,
  1973.

\bibitem[Pearl(2009)]{Pearl2009}
J.~Pearl.
\newblock \emph{{Causality: Models, Reasoning and Inference}}.
\newblock Cambridge University Press, 2009.

\bibitem[Pellet and Elisseeff(2008)]{Pellet2008a}
J.~P. Pellet and A.~Elisseeff.
\newblock {Using Markov blankets for causal structure learning}.
\newblock \emph{Journal of Machine Learning Research}, 9:\penalty0 1295--1342,
  2008.

\bibitem[Russakovsky et~al.(2014)Russakovsky, Deng, Su, Krause, Satheesh, Ma,
  Huang, Karpathy, Khosla, Bernstein, Berg, and Fei-Fei]{Russakovsky2014}
O.~Russakovsky, J.~Deng, H.~Su, J.~Krause, S.~Satheesh, S.~Ma, Z.~Huang,
  A.~Karpathy, A.~Khosla, M.~Bernstein, A.~C. Berg, and L.~Fei-Fei.
\newblock {ImageNet large scale visual recognition challenge}.
\newblock \emph{arXiv preprint arXiv:1409.0575}, 2014.

\bibitem[Settles and Craven(2008)]{Settles2008}
B.~Settles and M.~Craven.
\newblock {An analysis of active learning strategies for sequence labeling
  tasks}.
\newblock In \emph{Conference on Empirical Methods in Natural Langauge
  Processing}, pages 1070--1079, 2008.

\bibitem[Shalizi(2001)]{Shalizi2001a}
C.~R. Shalizi.
\newblock \emph{{Causal architecture, complexity and self-organization in the
  time series and cellular automata}}.
\newblock PhD thesis, University of Wisconsin at Madison, 2001.

\bibitem[Shalizi and Crutchfield(2001)]{Shalizi2001}
C.~R. Shalizi and J.~P. Crutchfield.
\newblock {Computational mechanics: Pattern and prediction, structure and
  simplicity}.
\newblock \emph{Journal of Statistical Physics}, 104\penalty0 (3-4):\penalty0
  817--879, 2001.

\bibitem[Spirtes and Scheines(2004)]{Spirtes2004a}
P.~Spirtes and R.~Scheines.
\newblock {Causal inference of ambiguous manipulations}.
\newblock \emph{Philosophy of Science}, 71\penalty0 (5):\penalty0 833--845,
  2004.

\bibitem[Spirtes et~al.(2000)Spirtes, Glymour, and Scheines]{Spirtes2000}
P.~Spirtes, C.~N. Glymour, and R.~Scheines.
\newblock \emph{{Causation, prediction, and search}}.
\newblock Massachusetts Institute of Technology, 2nd ed. edition, 2000.

\bibitem[Szegedy et~al.(2014)Szegedy, Zaremba, Sutskever, Bruna, Erhan,
  Goodfellow, and Fergus]{Fergus2014}
C.~Szegedy, W.~Zaremba, I.~Sutskever, J.~Bruna, D.~Erhan, I.~Goodfellow, and
  R.~Fergus.
\newblock {Intriguing properties of neural networks}.
\newblock In \emph{International Conference on Learning Representations}, 2014.

\bibitem[Vishwanathan(2010)]{Vishwanathan2010}
S.~V.~N. Vishwanathan.
\newblock {Graph kernels}.
\newblock \emph{Journal of Machine Learning Research}, 11:\penalty0 1201--1242,
  2010.

\bibitem[Zhang et~al.(2014)Zhang, Donahue, Girshick, and Darrell]{Zhang2014}
N.~Zhang, J.~Donahue, R.~Girshick, and T.~Darrell.
\newblock {Part-based R-CNNs for fine-grained category detection}.
\newblock In \emph{ECCV 2014}, pages 834--849, 2014.

\end{thebibliography}

\end{document}